\documentclass[a4paper]{llncs}
\usepackage{makeidx}  
\usepackage{graphicx}
\usepackage[utf8]{inputenc}        
\usepackage{times,amsmath,epsfig}
\usepackage{amssymb}
\usepackage{url}
\usepackage{array}
\usepackage[numbers]{natbib}

\begin{document}

\mainmatter              

\title{Top-down and Bottom-up Feature Combination for Multi-sensor Attentive Robots}

\author{Esther L. Colombini\inst{1} \and Alexandre S. Sim\~{o}es \inst{2} \and Carlos H. C. Ribeiro\inst{1}}

\authorrunning{Colombini et al.} 

\institute{Technological Institute of Aeronautics (ITA), S\~{a}o Jos\'{e} dos Campos, Brazil,\\
\email{esther, carlos@ita.br},
\and
S\~{a}o Paulo State University (UNESP), Sorocaba, Brazil,
\email{assimoes@sorocaba.unesp.br}}

\maketitle           

\begin{abstract}
The information available to robots in real tasks is widely distributed both in time and space, requiring the agent to search for relevant data. In humans, that face the same problem when sounds, images and smells are presented to their sensors in a daily scene, a natural system is applied: Attention. As vision plays an important role in our routine, most research regarding attention has involved this sensorial system and the same has been replicated to the robotics field. However, most of the robotics tasks nowadays do not rely only in visual data, that are still costly. To allow the use of attentive concepts with other robotics sensors that are usually used in tasks such as navigation, self-localization, searching and mapping, a generic attentional model has been previously proposed. In this work, feature mapping functions were designed to build feature maps to this attentive model from data from range scanner and sonar sensors. Experiments were performed in a high fidelity simulated robotics environment and results have demonstrated the capability of the model on dealing with both salient stimuli and goal-driven attention over multiple features extracted from multiple sensors.
\end{abstract}

\section{Introduction}

Attention has been mainly studied with visual experiments where the subject is looking to a scene that changes across time \cite{Navalpakkam_Itti_2005} \cite{Peters_Itti_2008}. In these models, the attentional system is usually restricted to its selective component in visual search tasks, focusing on the extraction of multiple features across the same sensor. However, most of the robotics tasks nowadays do not rely only in visual data, that are still costly. Although there are many computational models that apply attentive systems to robotics, they usually are restricted to two classes of systems: a) those that have complex biologically-based attentional visual systems \cite{Frintrop_2006}\cite{Yu_Mann_Gosine_2010} and b) those that have simpler attentional mechanisms with a variety of sensors \cite{Zeng_Weng_2007}. 

To allow the use of attentive concepts with other robotics sensors that are usually used in tasks such as navigation, self-localization, searching and mapping, a generic attentional model has been previously proposed in [removed]. These work present a computational attentional model that comprises the components of attention (orienting, selection and sustain) proposed in \cite{Posner1971}, inspired in the visual models of \cite{Navalpakkam_Itti_2005}\cite{Frintrop_2006}. 

Based on the model presented in previous work, this paper focuses on describing how exogenous and endogenous attentional processes can be modeled to work with a multi-dimensional multi-sensorial environment. A set of proper feature mapping functions were proposed. Experiments were conducted in a search and rescue high fidelity simulator with range scanner and sonars sensors embedded in mobile robots. 

The rest of this paper is organized as follows. Section \ref{sec:model} presents the general structure of the previously proposed model whereas section \ref{sec:methods} details the feature extraction process, the environment set-up and the proposed experiments. Section \ref{sec:full} shows the results achieved in order to validate the proposed features and, at last, section \ref{sec:conclusion} presents discussions regarding the results, concluding with assessments to direct further investigation.

\section{Attentional Model} \label{sec:model}

\begin{figure*}[htb]
\centering
\includegraphics [width=13cm] {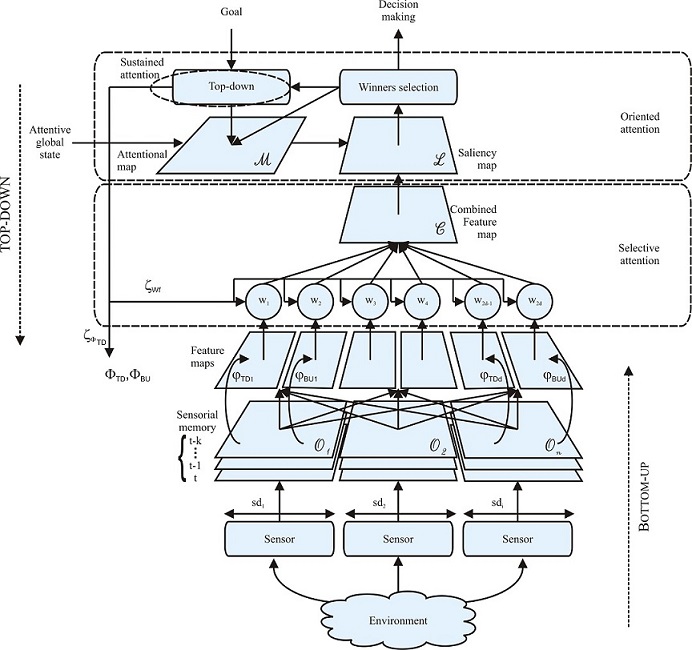}
\caption{Proposed Attentional Framework}
\label{fig:colombini}
\end{figure*}

Figure \ref{fig:colombini} shows the attentional framework proposed in previous work \cite{Colombini2013}, that represents an extension of the model proposed in \cite{Colombini2012}, accommodating top-down influences. In this model, multiple \textbf{Sensors} (${s_{i}}$) have their data stored in the \textbf{Sensorial Memory }(${O}_{i}$) during a time window and they can individually or grouped contribute with information to build \textbf{Feature Maps} (${F}_d$). These feature maps are built through \textbf{Feature Dimension Functions} ($\phi$) that are applied over the \textbf{Observation Spaces} (${O}_{i}$).  The feature dimension functions can work both from the bottom-up ($\phi_{BU}$) and the top-down ($\phi_{TD}$) perspective. The corresponding feature maps are weighted generating the \textbf{Combined Feature Map} ($\mathcal{C}$).

Finally, the \textbf{Attentional Map} ($\mathcal{M}$), that represents the current attentional status of the system, and the combined feature map produce a \textbf{Saliency Map} ($\mathcal{L}$). This saliency map is used by a \textbf{Winners Selection} component (modeled here as a winner-takes-all element) to elicit a sub-representation of the perceived environment, modulated by the attentional process, to a decision maker and long-term storing units. 

From a bottom-up perspective, these feature maps provide information that represent saliences in the environment that would require attention and that, if attended, will enhance the corresponding region and its surrounding in the attentional map ($\mathcal{M}$) at time ($t$) for another period ($t + y$). After that, the inhibition of return (IOR) \cite{Klein_2000} effect will suppress the previously enhanced region for another period of time. The lateral excitatory and inhibitory influences decay with time based on the curve adopted to model neuronal activity integration \cite{Parkhurst2002a}.

The top-down pathway allows voluntary attention to, depending on the system goal(s), promote a region in the attentional map ($\mathcal{M}$) in two manners: i) through a function ($\phi_{TD}$) that considers the goal to build the corresponding feature map based on the observations, or ii) by adjusting the weights ($\mathcal{W}_f$) that define the contribution of each feature dimension. If voluntary attention enhances a region or feature of interest, the model takes into consideration the work of \cite{Ling2006} to define the top-down contribution over the attentional map along time. In Ling's work, vigilance over a region produces an adaptation that causes inhibition in the attended region after long exposure, requiring more signal strength to regain attention for a certain period. 

In the process chronometry, as investigated by \cite{Castro2004}, the course of bottom-up attention starts about 150 ms after the stimulus is presented and it lasts for another 150 ms, when it starts to impair perception for another 300 ms (inhibition of return). As for top-down course, the enhancement starts about 200 ms post-stimulus on set and it lasts as long as the attention is driven towards it, stopping almost immediately as the stimulus fades away without post impairment if vigilance is not reached (6s). 

In fact, in the way the system has been proposed, bottom-up and top-down influences can be summed up when aligned (due to the feature map combination) or compete when looking for different characteristics. The whole attentional system can also be modulated by a \textbf{Global Attentive State} (adapted from \cite{Gazzaniga2002}). A formal definition of the proposed attentional model can be found in \cite{Colombini2013}.

\section{Material and Methods}
\label{sec:methods}

This section presents the feature extracting process proposed for the model depicted in Section \ref{sec:model} in a robotics task. 

\subsection{Simulated Environment}

\begin{figure*}[!htbp]
\begin{center}
		\begin{tabular} {cc}
		\includegraphics[width=5.0cm]{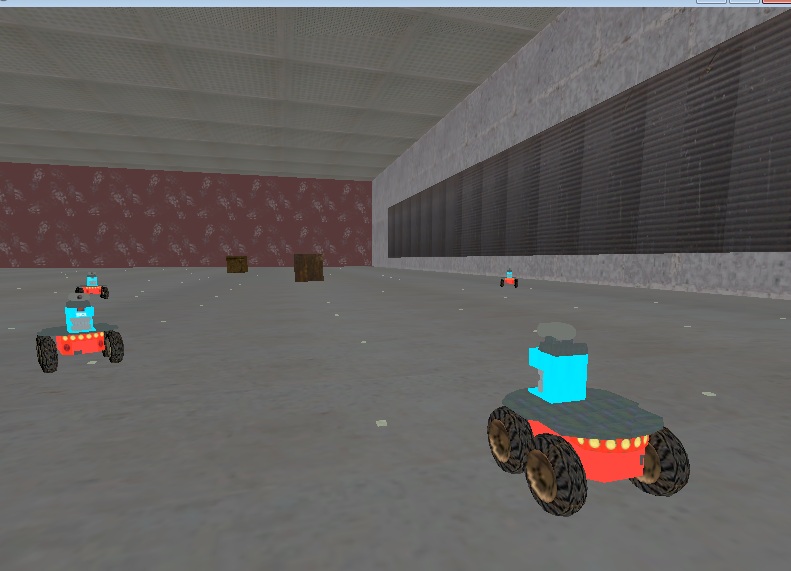} &
		\includegraphics[width=7.5cm]{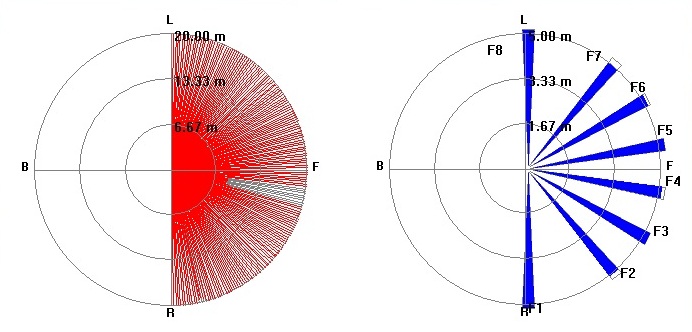} \\
		(a) & (b) \\
		\end{tabular}
\end{center}
\caption{USARSim Simulation Environment with Robots Deployed. a) Simulated Robot. b) Sensors distribution (left: range scanner; right: sonars)}
\label{fig:robots}
   \end{figure*}

The experiments presented in this work were developed based on the USARSim platform \cite{Wang2006}, a high fidelity simulator used for multi-robot coordination. Commercially available robots and sensors are modeled in this simulator, including the ground wheeled Pioneer P2AT robot used in this work.  The simulated robot is equipped with 16 sonar sensors with 15cm-5m range, 8 of them disposed in the robot frontal $180^0$ and the other 8 in the rear, and a range scanner with $180^0$ aperture and saturation at 20m. A 5\% noise was adopted in all sensors. A sample test scenario with four robots deployed can be seen in Figure \ref{fig:robots}a whereas the sensor distribution is represented in \ref{fig:robots}b.

\subsection{Data Observation}
\label{sec:data}

In order to obtain the feature maps for each dimension that will be used to compute saliences and to perform top-down attention, a temporal window of data was stored in the sensorial memory regarding the 8-front sonar readings (denoted by $sonar$) and the $180^0$  range scanner measurements (denoted by $range$). The sensors were aligned in such a way that they will capture data from the environment around the frontal $180^0$ of the attentive robot. The observation spaces are defined as:

\begin{itemize}
	\item $ O_{1}$ represents the observation space for sonar readings, defined by: $ o_{1_{n_{t}}} = sonar_{n_{t}}$ with $n \in [1,8]$
	\item $ O_{2}$ represents the observation space for range scanner readings, defined by: $ o_{2_{n_{t}}} = range_{n_{t}}$ with $n \in [1,180]$
\end{itemize}

\subsection{Feature Extraction}

To evaluate the feasibility of combining different feature dimensions provided by multiple sensors under a robotics task, four dimensions usually relevant in robotics environments were assumed:

\begin{itemize}
	\item \textbf{Motion}: represents the intensity of motion detected in the environment;
	\item \textbf{Direction}: represents the direction of moving objects in the environment relative to the attentive agent.
	\item \textbf{Distance}: represents the distance between obstacles (static or moving) and the attentive agents;
\end{itemize}

To obtain these features, $3$ bottom-up and $3$ top-down feature mapping functions were designed based on the observation maps defined in Section \ref{sec:data} and detailed in section \ref{sec:tdfeatures}. Moreover, to understand the influence of combining multiple feature dimensions generated by sensors with different capabilities, some feature maps were constructed from data of multiple observation spaces.

\subsubsection{Bottom-up feature mapping functions.}
\label{sec:bufeatures}

When defining feature mapping functions that aim to identify salient stimuli that are relevant in a robotics task based on a non-visual environment, a transfer of domain is required. Whereas most of the computational models that deal with image create many maps for one dimension, one for each discrete value of the continuous set,  in this work, a different approach is adopted. Here, the feature map for the dimension contains continuous values representing the discrepancy level among its observations. First, we introduce the Motion feature mapping function.

\textbf{$\mathcal{F}_1$: Motion}
\label{sec:motion}

This feature extracts regions with strong speed contrast by estimating the speed targets/distractors move in the scene. Assuming that each moving obstacle in the environment moves with constant speed, it computes $\phi{_{BU_1}} = z (O_{1})$ where $z (O_{1})$ is given by:

	\begin{equation}
		f_{1_{n_t}} = |\frac{\Delta s_{n}}{\Delta t}|
	\end{equation}
	\label{eq:speed}

,where $\Delta t$ is the time variation returned by the simulator between two sonar readings and $\Delta s =  o_{1_{n_{t}}} - o_{1_{n_{t-1}}} $. To establish the level of contrast among the set elements, each motion feature element was update according to:

	\begin{equation}
		f_{1_{n_t}} = |f_{1_{n_t}} - \frac{\sum\limits_{m=1}^{n} f_{1_{m_t}}}{n}|
	\end{equation}
	
Finally, each feature element was normalized by dividing the elements by the maximum allowed speed for moving objects in the the tested environment (MAX-SPEED). 

\textbf{$\mathcal{F}_2$: Direction}
\label{sec:direction}

This feature extracts information regarding the direction of moving objects (targets/distractors) in the environment. The attentive robot is used as reference and three directions were considered: a) towards the attentive robot; b) away from the attentive robot; c) not changing direction. The function $\phi{_{BU_2}} = z (O_{1}) \circ z (O_{2})$ where $z (O_{1})$ is given by:

	\begin{equation}
		tf_{1_{n_t}} =  
		\left\{ 
			\begin{array}{lll}
				1 & if & \frac{\Delta s_{n}}{\Delta t} > 0\\ 
        -1 & if & \frac{\Delta s_{n}}{\Delta t} < 0\\ 
				0 & & otherwise.
			\end{array} \right.
	\end{equation}
	\label{eq:direction}

,where $\Delta t$ and $\Delta s$ are the same as Equation \ref{eq:speed} and  $tf_{1_{n_t}}$ is a temporary feature map with dimension $n$, with $n \in [1,8]$ . Then, to measure the discrepancy level, each $tf_{1_{n_t}}$ is updated according to:

	\begin{equation}
		tf_{1_{n_t}} =  \frac{n}{count(tf_{1_{n_t}})}  
	\end{equation}
	\label{eq:distance1_1}

where $count(tf_{1_{n_t}})$ is a function the determines the number of occurrences of this feature value in the temporary feature dimension. 

To establish whether there are benefits on using sensors with different capabilities (range, material sensibility, etc) to generate the same feature map,  $z (O_{2})$ is represented in the same way as Equation \ref{eq:direction}, with $\Delta t$ the time variation returned by the simulator between two range scanner readings and $\Delta s =  o_{2_{n_{t}}} - o_{2_{n_{t-1}}} $, $n \in [1,180]$, as in $O_{2}$. Then, $tf_{2_{n_t}} = n/count(tf_{2_{n_t}})$. The composite function $\phi{_{BU_2}} = z (O_{1}) \circ z (O_{2})$ uses both temporary feature spaces to determine each $f_{2_{n_t}}$ of feature map $\mathcal{F}_2$. The composition is carried out by reducing the dimension of $z (O_{2})$ in order to fit its data to their relative in $z (O_{1})$. 
	
\textbf{$\mathcal{F}_3$: Distance}
\label{sec:distance}

$\phi{_{BU_3}}$ is responsible for extracting information regarding the disposition of elements around the attentive agent. $\phi{_{BU_3}} = z (O_{2})$ where $z (O_{2})$ is given by:

	\begin{equation}
		f_{3_{n_t}} = |o_{2_{n_{t}}} - \frac{\sum\limits_{m=1}^{n} o_{2_{m_{t}}}}{n}| / mr
	\end{equation}
	\label{eq:distance}

, where $mr$ is the saturation value of the range scanner.

\subsubsection{Top-down feature mapping functions.}
\label{sec:tdfeatures}

Although bottom-up driven attention plays an important role in our daily routine by capturing one's attention to situations that pop-out in the scene, most of our attention is driven by top-down influence. In order to drive attention to regions of interest or features of interest over the observation space defined in section \ref{sec:data}, functions that map goals to the feature space are defined. The $3$ top-down mapping functions are: GoalSpeed, GoalDirection and GoalDistance.

\textbf{$\mathcal{F}_4$: GoalSpeed}

Consider the value $f_{1_{n_t}}$ found by equation \ref{eq:speed} as the module of the speed of each target/distractor relative to the attentive agent in time $t$. Then, each $f_{4_{n_t}}$ value of the goal speed feature map will be computed as show in Table \ref{table:topdownSpeed} for the following mapped goals:

\begin{enumerate}
	\item EQUAL($desiredSpeed$): to find elements with speed equal to $desiredSpeed$;
	\item BETWEEN($desiredSpeed$,$desiredSpeed + \Delta Speed$): to find elements with speed between $desiredSpeed$ and $desiredSpeed + \Delta Speed$;
	\item GREATER($desiredSpeed$): to find elements with speed greater than the $desiredSpeed$;
	\item SMALLER($desiredSpeed$): to find elements with speed less than the $desiredSpeed$.
\end{enumerate}

\begin{table}[!htbp]
\caption{Pseudo-code for the top-down goal mapping function $\phi_{TD_1}$}\label{table:topdownSpeed}
\begin{center}
\begin{tabular}{|l|}
\hline
\hspace{0cm} if \textit{goal} is EQUAL and $f_{1_{n_t}} == desiredSpeed$ then $f_{4_{n_t}} = 1$ \\
\hspace{0cm} if \textit{goal} is EQUAL and $f_{1_{n_t}} != desiredSpeed$ then $f_{4_{n_t}} = f_{1_{n_t}}$/MAXSPEED \\
\hspace{0cm} if \textit{goal} is BETWEEN and $f_{1_{n_t}} >= desiredSpeed$ and $f_{1_{n_t}} <= desiredSpeed$ then $f_{4_{n_t}} = 1$ \\
\hspace{0cm} if \textit{goal} is BETWEEN  and $f_{1_{n_t}} < desiredSpeed$ then $f_{4_{n_t}} = f_{1_{n_t}}$/$desiredSpeed$ \\
\hspace{0cm} if \textit{goal} is BETWEEN  and $f_{1_{n_t}} > desiredSpeed + \Delta Speed$ then \\ 
\hspace{0.4cm} $f_{4_{n_t}} = 1 - (f_{1_{n_t}}-$ $desiredSpeed+\Delta Speed$/(MAXSPEED-$ desiredSpeed+\Delta Speed$))\\
\hspace{0cm} if \textit{goal} is GREATER and $f_{1_{n_t}} > desiredSpeed$ then $f_{4_{n_t}} = 1$ \\
\hspace{0cm} if \textit{goal} is GREATER and $f_{1_{n_t}} <= desiredSpeed$ then $f_{4_{n_t}} = f_{1_{n_t}}$/$desiredSpeed$ \\
\hspace{0cm} if \textit{goal} is SMALLER and $f_{1_{n_t}} < desiredSpeed$ then $f_{4_{n_t}} = 1$ \\
\hspace{0cm} if \textit{goal} is SMALLER and $f_{1_{n_t}} >= desiredSpeed$ then \\
\hspace{0.4cm}  $f_{4_{n_t}} = 1 - (f_{1_{n_t}}- desiredSpeed$/(MAXSPEED-$desiredSpeed$))\\
\hline
\end{tabular}
\end{center}
\end{table}

Although this mapping has been defined manually, it could be found automatically, as proposed by \cite{Frintrop_2006}\cite{Frintrop2006}. Although goals can set the influence of features of interest to the maximum level, this process can be modulated by the weight of the specific feature map, as described in \ref{fig:colombini}. 

\textbf{$\mathcal{F}_5$: GoalDirection}

Consider the value $f_{2_{n_t}}$ found by Equation \ref{eq:direction} as the module of the speed of each target/distractor relative to the attentive agent in time $t$. Then, each $f_{5_{n_t}}$ value of the top-down feature map will be computed as show in Table \ref{table:topdownDirection} for the following mapped goal:

\begin{enumerate}
	\item EQUAL($desiredDirection$): to find elements with the same direction of movement as $desiredDirection$;
\end{enumerate}

\begin{table}[!htbp]
\caption{Pseudo-code for the top-down goal mapping function $\phi_{TD_2}$}\label{table:topdownDirection}
\begin{center}
\begin{tabular}{|l|}
\hline
\hspace{0cm} if \textit{goal} is EQUAL and $f_{2_{n_t}} == desiredDirection$ then $f_{5_{n_t}} = 1$ \\
\hspace{0cm} if \textit{goal} is EQUAL and $f_{2_{n_t}} != desiredDirection$ then $f_{5_{n_t}} = 0$ \\
\hline
\end{tabular}
\end{center}
\end{table}

\textbf{$\mathcal{F}_6$: GoalDistance}

Consider the value $f_{3_{n_t}}$ found by equation \ref{eq:distance} as the disposition of elements around the attentive agent in time $t$. Then, each $f_{6_{n_t}}$ value of the top-down feature map will be computed as show in Table \ref{table:topdownDistance} for the following mapped goals:

\begin{enumerate}
	\item EQUAL($desiredDistance$): to find elements with speed equal to $desiredSpeed$;
	\item BETWEEN($desiredDistance$,$desiredDistance + \Delta Distance$): to find elements with speed between $desiredDistance$ and $\textbf{desiredSpeed} + \Delta Distance$;
	\item GREATER($desiredDistance$): to find elements with speed greater than the $desiredDistance$;
	\item SMALLER($desiredDistance$): to find elements with speed less than the $desiredDistance$.
\end{enumerate}

\begin{table}[!htbp]
\caption{Pseudo-code for the top-down goal mapping function $\phi_{TD_3}$}\label{table:topdownDistance}
\begin{center}
\begin{tabular}{|l|}
\hline
\hspace{0cm} if \textit{goal} is EQUAL and $f_{3_{n_t}} == desiredDistance$ then $f_{6_{n_t}} = 1$ \\
\hspace{0cm} if \textit{goal} is EQUAL and $f_{3_{n_t}} != desiredDistance$ then $f_{6_{n_t}} = f_{3_{n_t}}$/MAXDISTANCE \\
\hspace{0cm} if \textit{goal} is BETWEEN and $f_{3_{n_t}} >= desiredDistance$ and $f_{3_{n_t}} <= desiredDistance$ then $f_{6_{n_t}} = 1$ \\
\hspace{0cm} if \textit{goal} is BETWEEN  and $f_{3_{n_t}} < desiredDistance$ then $f_{6_{n_t}} = f_{3_{n_t}}$/$desiredDistance$ \\
\hspace{0cm} if \textit{goal} is BETWEEN  and $f_{3_{n_t}} > desiredDistance + \Delta Distance$ then \\ 
\hspace{0.4cm} $f_{6_{n_t}} = 1 - (f_{3_{n_t}}-$ $desiredDistance+\Delta Distance$/(MAXDISTANCE-$ desiredDistance+\Delta Distance$))\\
\hspace{0cm} if \textit{goal} is GREATER and $f_{6_{n_t}} > desiredDistance$ then $f_{6_{n_t}} = 1$ \\
\hspace{0cm} if \textit{goal} is GREATER and $f_{6_{n_t}} <= desiredDistance$ then $f_{6_{n_t}} = f_{3_{n_t}}$/$desiredDistance$ \\
\hspace{0cm} if \textit{goal} is SMALLER and $f_{6_{n_t}} < desiredDistance$ then $f_{6_{n_t}} = 1$ \\
\hspace{0cm} if \textit{goal} is SMALLER and $f_{6_{n_t}} >= desiredDistance$ then \\
\hspace{0.4cm}  $f_{6_{n_t}} = 1 - (f_{3_{n_t}}- desiredDistance$/(MAXSPEED-$desiredDistance$))\\
\hline
\end{tabular}
\end{center}
\end{table}

%
%
%
	%
%

\section{Experiments}

In this work, all experiments performed consider only one immobile attentive robot deployed in the environment. A goal (from Section \ref{sec:tdfeatures}) can be assigned to the this robot and it can change during the same experiment. A winner-takes-all algorithm was adopted to define the most salient feature/region. A threshold was established as the minimum value for firing a winner. 

Two sets of experiments were employed. The first set aims at validating the use of single features extracted from single or multiple observation spaces (Section \ref{sec:singleValidation} - Experiments 01-02). The second set of experiments is performed in order to evaluate the conjunction of the multiple features proposed. It also evaluates the ability of the model on handling simultaneous bottom-up and top-down attention (Section \ref{sec:multipleValidation} - Experiments 03-05).

\subsection{Single feature validation}
\label{sec:singleValidation}

\begin{itemize}
	\item \textbf{Exp 01} (Single Dimension): in this experiment, one robot running the attentional algorithm of Section \ref{sec:model} was deployed in the environment. Another robot was deployed 3 meters far from the first robot and it moves towards the attentive agent for 0.4m. Then, it moves backwards until it is not reached by the sonar sensors. In this experiment, only $F_1$, described in section \ref{sec:motion}, was applied.
	\item \textbf{Exp 02} (Multiple Dimensions):  in this experiment, one attentive robot was deployed in the environment. Another robot was deployed 3 meters far from the first robot and it moves towards the attentive agent until it passes it and stay out of the sensor's reach. In this experiment, only  $F_3$, described in section \ref{sec:distance}, was applied.
\end{itemize}

\subsection{Multiple features validation}
\label{sec:multipleValidation}

\begin{itemize}
	\item \textbf{Exp 03}: In this experiment, one robot running the attentional algorithm was deployed in the environment. Other two non-attentive robots were deployed about 8 meters far from the first robot. The two robots move at same speed, but they start moving at different moments. They both get close to the attentive robot and then come backwards. There is a wall in the robot side and two static obstacles in the far front. In this experiment, two feature dimensions were employed:  $\mathcal{F}_1$, described in section \ref{sec:motion}, and  $\mathcal{F}_3$, described in \ref{sec:distance}.	
	\item \textbf{Exp 04}: In this experiment, one attentive robot was deployed in the environment. Another robot was deployed outside the attentive robot sensors range, and it moved in the direction of the first one and then backwards with approximate speed of 0.7m/s. Features  $\mathcal{F}_1$ and  $\mathcal{F}_4$ were employed. For $\mathcal{F}_4$, the top-down goal was defined as BETWEEN(0.5,0.9), in m/s.
	\item \textbf{Exp 05}: In this experiment, one attentive robot was deployed in the environment. Another two robots were deployed outside the attentive robot sensors range, and they moved in the direction of the first one and then backwards with approximate speed of 0.7m/s and 0.6m/s respectively. Features $\mathcal{F}_1$ and $\mathcal{F}_4$ were employed. For $\mathcal{F}_4$, the top-down goal was defined as GREATER(0.6), in m/s. 
\end{itemize}

\section{Results and Discussion}
\label{sec:full}

\subsection{Experiment 01}

\begin{figure}[!tb]
\begin{center}
		\begin{tabular} {cc}
		\includegraphics[height=3cm]{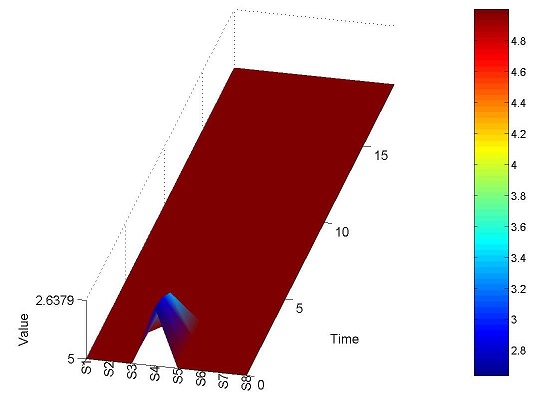} &
		\includegraphics[height=3cm]{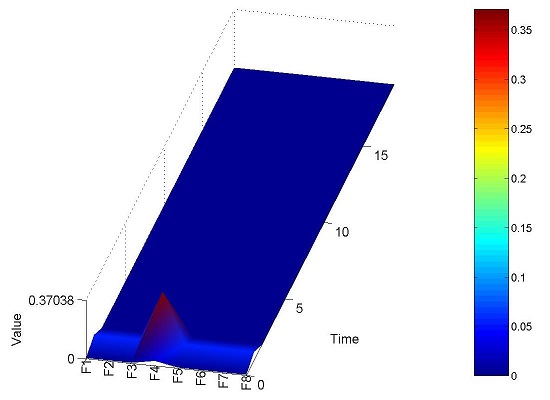} \\
		(a) & (b) \\
		\includegraphics[height=3cm]{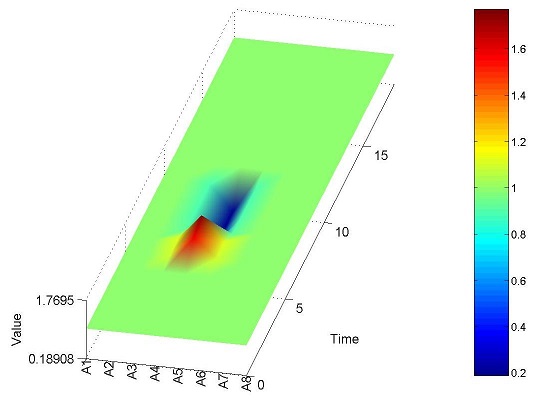} &
		\includegraphics[height=3cm]{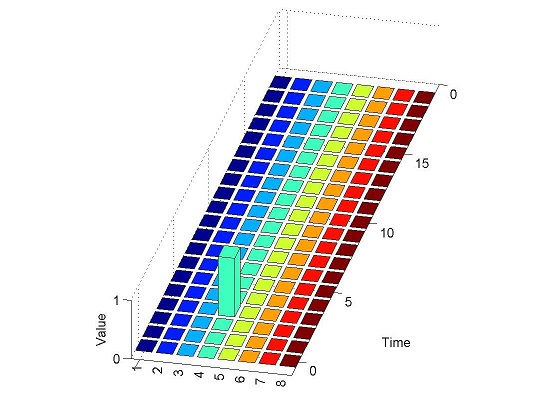} \\
		(c) & (d) \\
		\end{tabular}
\end{center}
\caption{Experiment 01: a) $O_{1}$ b) $\mathcal{F}_1$ c) $\mathcal{M}$ d) WTA.}
\label{exp1}
   \end{figure}

Results are shown in Figure \ref{exp1}. As it can be seen, the observation relative to $s4$ (a) correctly captured the moving robot and it generated a salient feature for the motion dimension (b), once the other robot was moving with significant speed. A saliency map was generated over this feature dimension and a winner was chosen (d) indicating that there was an exogenous salient stimulus. A relevant stimulus in the feature dimension occurred in $t$ = 1, generating the expected course of excitatory and inhibitory stimulation in the attentional map (c) after the pre-activation delay. No second relevant stimulus occurred later as the target moved away from the sensor range and a static environment with no salient motion was in place. 

\subsection{Experiment 02}

\begin{figure*}[!htb]
\begin{center}
		\begin{tabular} {ccc}
		\includegraphics[height=3cm]{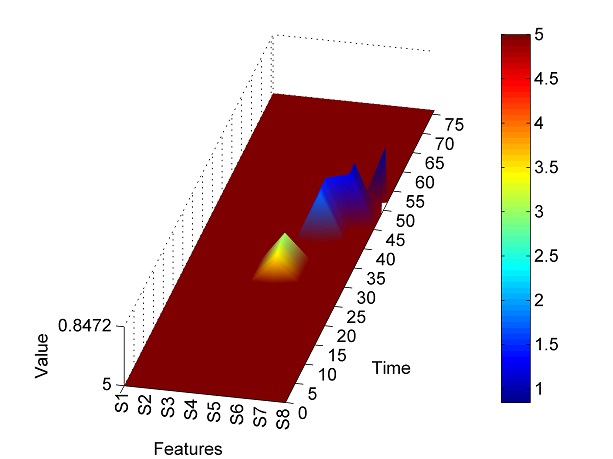} &
		\includegraphics[height=3cm]{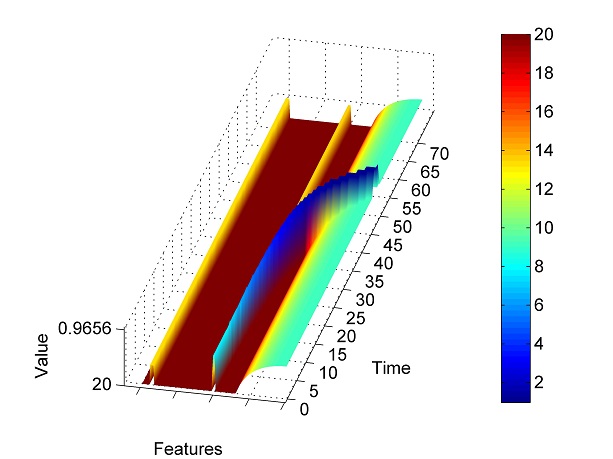} &
		\includegraphics[height=3cm]{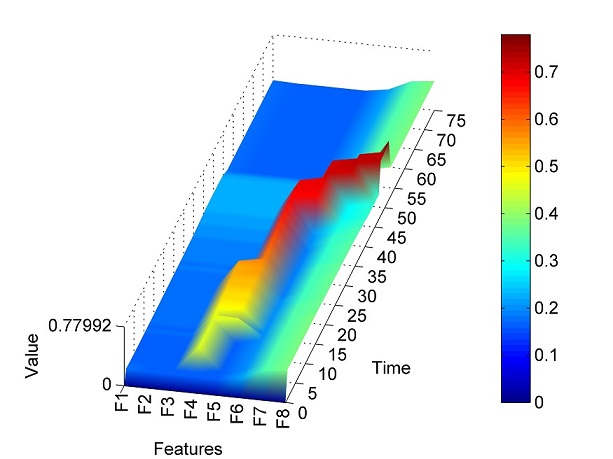}\\
		(a) & (b) & (c) \\
		\includegraphics[height=3cm]{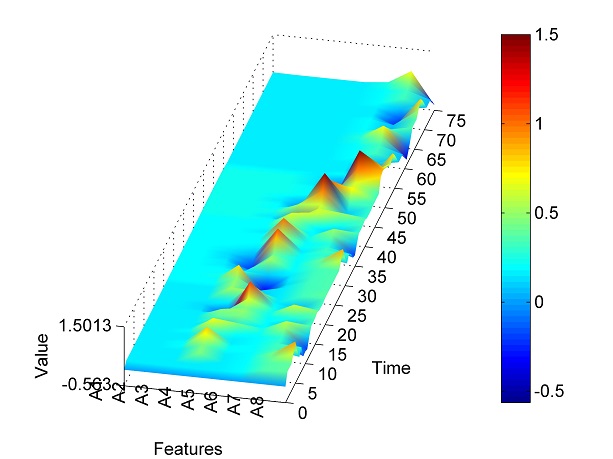}  &
		\includegraphics[height=3cm]{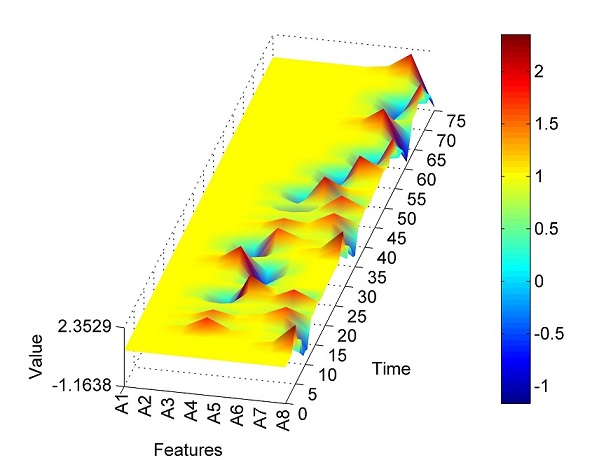} &
		\includegraphics[height=3cm]{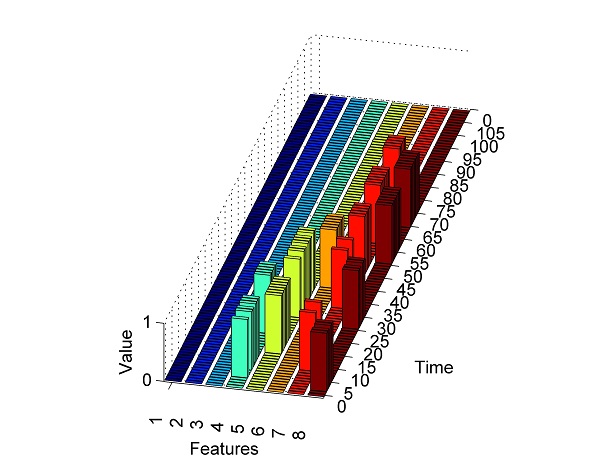}  \\
		(d) & (e) & (f) \\
		\end{tabular}
\end{center}
\caption{Experiment 02: a) $O_{1}$ b) $O_{2}$ c) $\mathcal{F}_3$ d) $\mathcal{L}$ e) $\mathcal{M}$ f) WTA.}
\label{exp2}
   \end{figure*}

Figure \ref{exp2} depicts the results for experiment 2, with (a) representing the sonar readings over time, where it is possible to see the moving robot being captured by the sonar readings in time $30$. Figure \ref{exp2} (b) represents the range scanner readings. It is possible to verify that this sensor was capturing obstacles since the beginning of the experiment (fixed walls) that were out of the range of the sonars. As it can be observed, the distance dimension (c) was able to capture information provided by the sonar sensors and from the range scanner. As the bottom-up distance feature  $\mathcal{F}_3$  detects discrepancies relative to obstacles distances in the environment, at the beginning, the range scanner understood that there was a relevant discrepancy regarding the fixed obstacles and the empty space, firing the expected course of excitatory and inhibitory stimulation took place in the attentional map (e) after the pre-activation delay, at time $t=5$. Yet, as a target robot started moving towards the attentive robot, around $t=30$ it entered in the sonar range, increasing the distance discrepancy in the environment and generating new relevant stimulus that amplified the excitatory component in the current map. This shows that the attentional model did promote attended locations for a certain period after a relevant stimulus is presented. The inhibition model prevented a new driving of the attentional focus to a recently previous attended location, showing that IOR effect is a natural mechanism that promotes exploration. It is worth to mention the ability of two sensor with different ranges work jointly to detect stimuli. In this experiment two different sensors were used to build the same feature dimension.

\subsection{Experiment 03}

\begin{figure*}[!htb]
\begin{center}
		\begin{tabular} {ccc}
		\includegraphics[height=3cm]{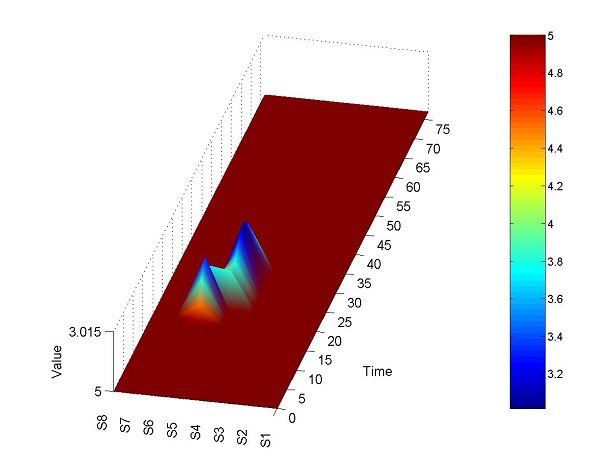} &
		\includegraphics[height=3cm]{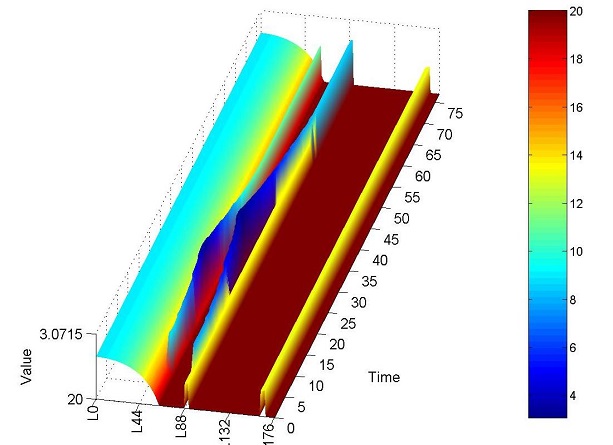} &
		\includegraphics[height=3cm]{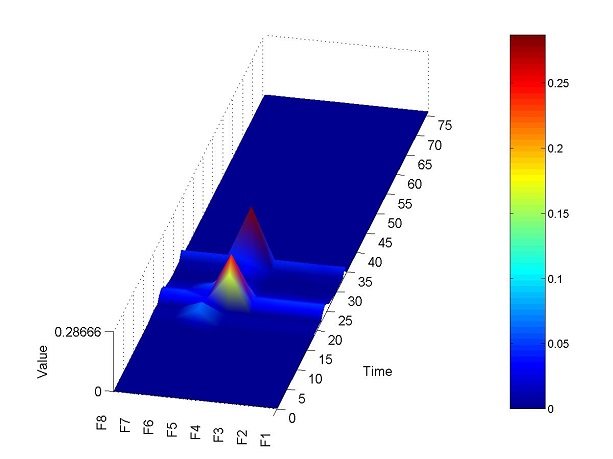} \\
		(a) & (b) & (c)\\
		\includegraphics[height=3cm]{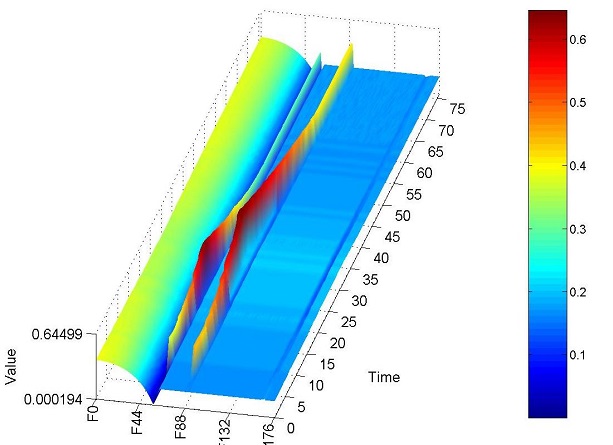}  &
		\includegraphics[height=3cm]{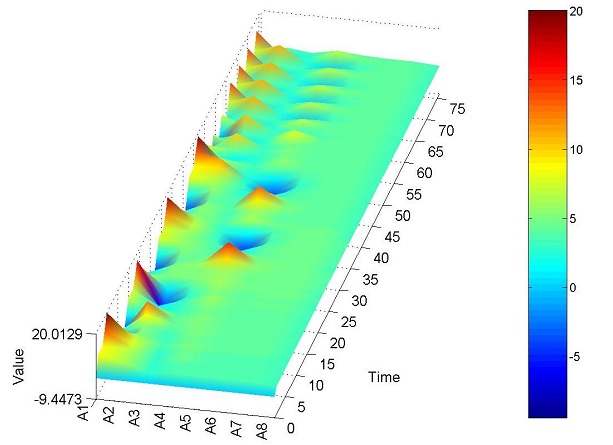} &
		\includegraphics[height=3cm]{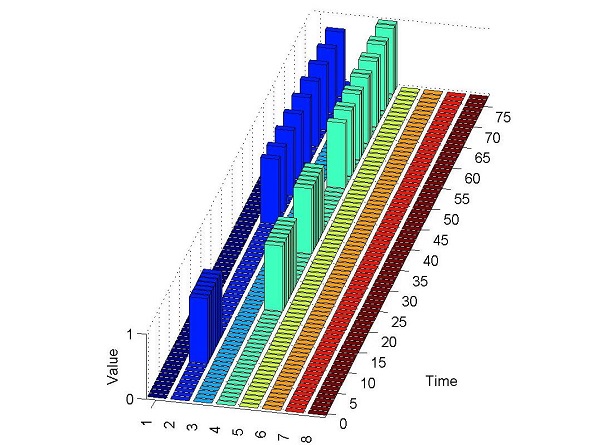}  \\
		(d) & (e) & (f) \\
		\end{tabular}
\end{center}
\caption{Experiment 03: a) $O_{1}$ b) $O_{2}$ c)  $\mathcal{F}_1$ d)  $\mathcal{F}_3$ e) $\mathcal{L}$ f) WTA.}
\label{exp12}
   \end{figure*}

In this experiment, presented in Figure \ref{exp12}, two different feature dimensions were build from mappings of two distinct sources: laser data (b) to measure distance (d) and sonar data (a) to detect motion (c). At the beginning of the experiment, as no motion was detected by the sonars, the distance feature was predominant, generating all salience in the left side of Figure (e), with the correspondent winners in (f). However, once the motion feature initiated its influence ($t=25$) and it was added to the enhancement of the distance feature towards this moving robot, greater was the final result. Until $t = 40$, the only time this region lost attendance was during refractory period. It is important to notice that the influence of the distance feature is higher in the context due to the way it was modeled.  It it worthy to mention that one could give more relevance to a feature dimension in this architecture only by modifying the weighted sum to build up the combined map that originated the salience (e).

\subsection{Experiment 04}

\begin{figure}[!hbnt]
\begin{center}
		\begin{tabular} {ccc}
		\includegraphics[height=3cm]{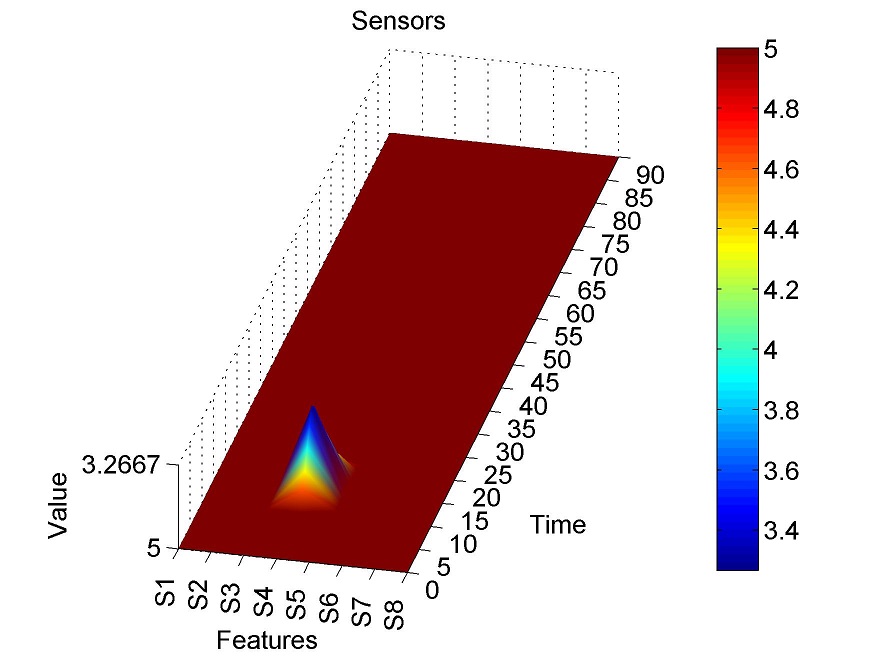} &
		\includegraphics[height=3cm]{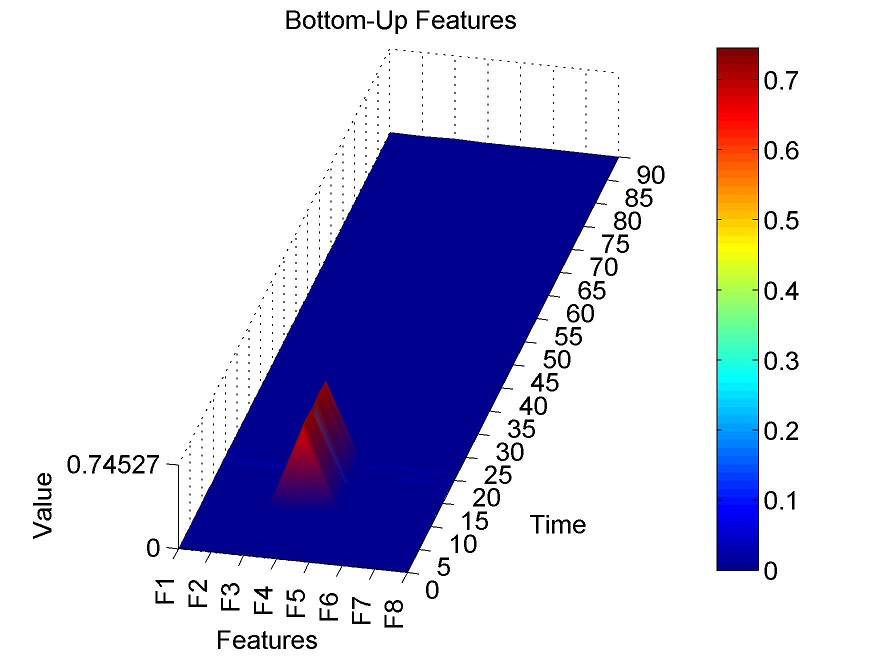} &
		\includegraphics[height=3cm]{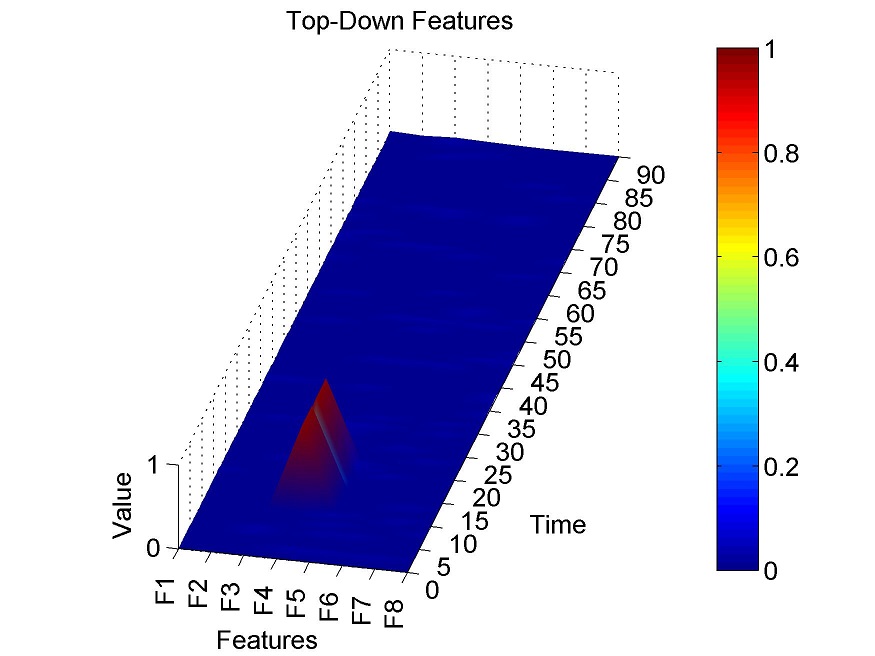}\\
		(a) & (b) & (c) \\
		\includegraphics[height=3cm]{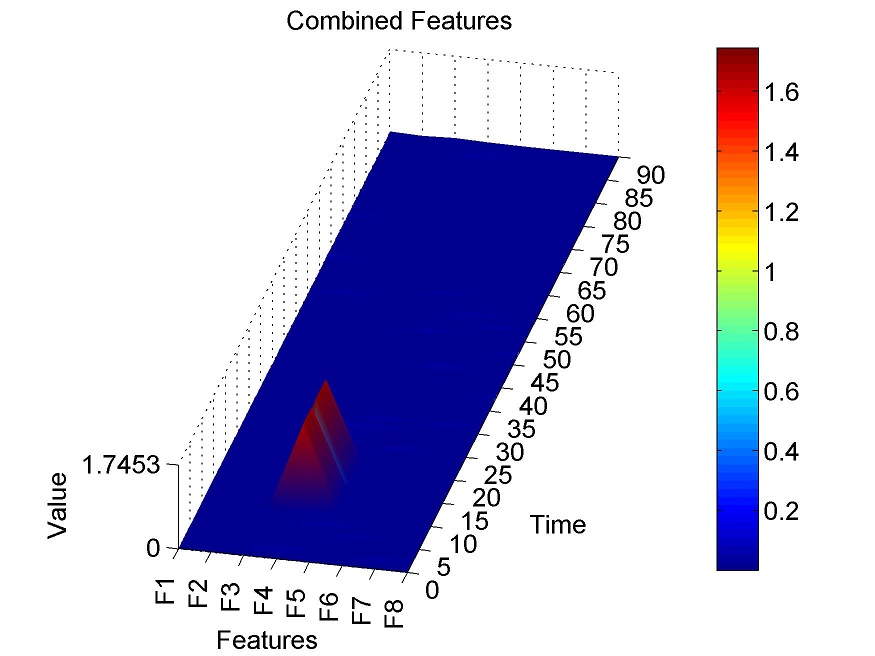}  &
		\includegraphics[height=3cm]{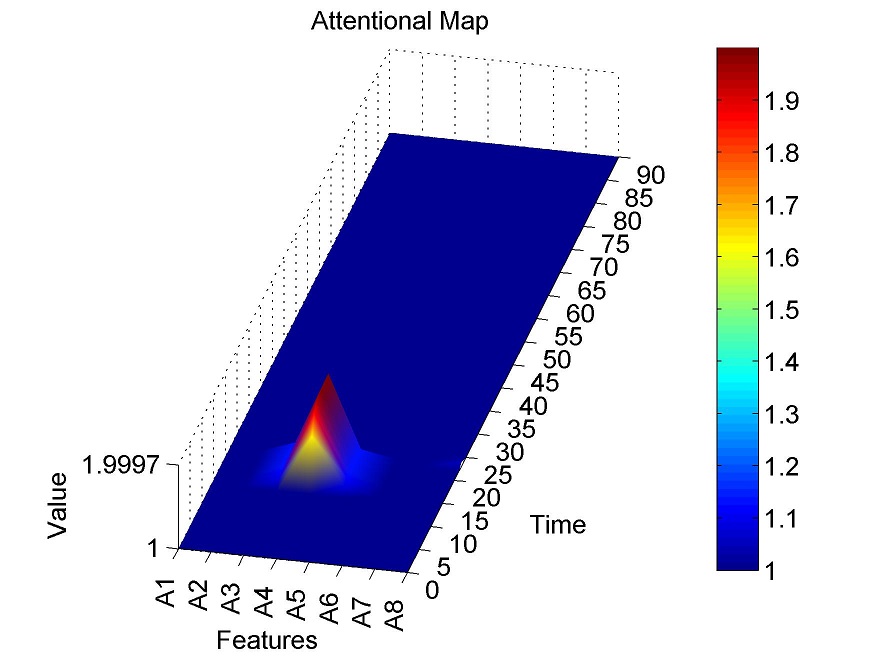} &
		\includegraphics[height=3cm] {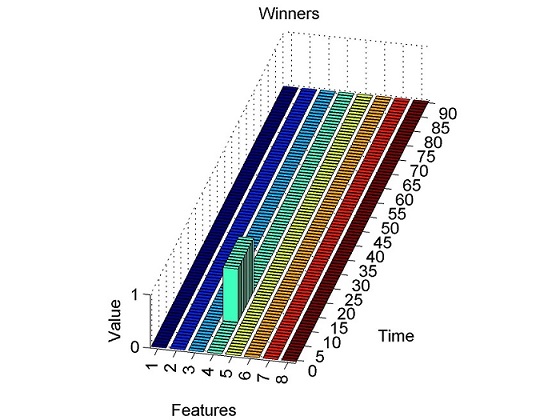}\\ 
		(d) & (e) & (f) \\
		\end{tabular}
\end{center}
\caption{Experiment 04: a) $\mathcal{O}_1$ b) $\mathcal{F}_1$ c) $\mathcal{F}_4$ d) $\mathcal{C}$ e) $\mathcal{M}$ f) WTA.}
\label{exp04}
   \end{figure}

Figure \ref{exp04} presents the result for this experiment. As it can be observed, the sensor 4 captured correctly the other robot and it generated a salient feature for the speed dimension in the bottom-up and corresponding signal for the goal speed top-down map. While there was fulfillment of the goal, the top-down mechanism enhanced the region activity. Moreover, as both top-down and bottom-up mechanisms (one due to goal and the other due to discrepancy) were interested in the same feature, the final salience was even stronger. However, as the activation was generated by a top-down source, that had greater importance, no inhibition of return effect occurred due to the fact that, top-down attentional mechanisms enhance the region/feature of interest as long as they require, without impairing future stimulus. Also, the activated region lost the focus almost immediately when the goal was not fulfilled. 

\subsection{Experiment 05}

\begin{figure}[!hbnt]
\begin{center}
		\begin{tabular} {ccc}
		\includegraphics[height=3cm]{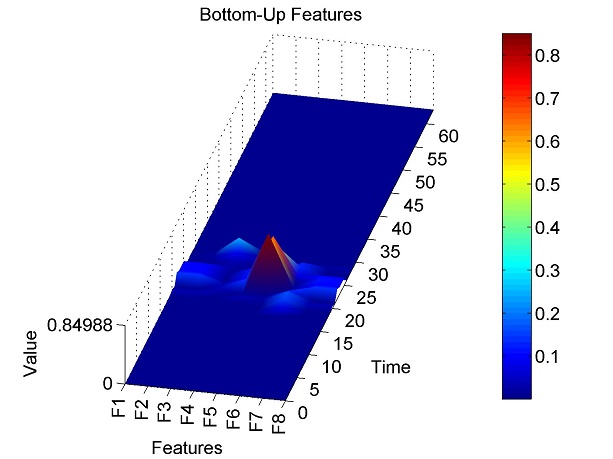} &
		\includegraphics[height=3cm]{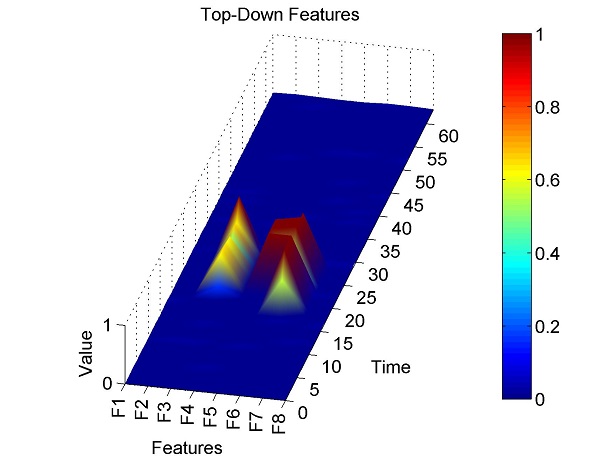} &
		\includegraphics[height=3cm]{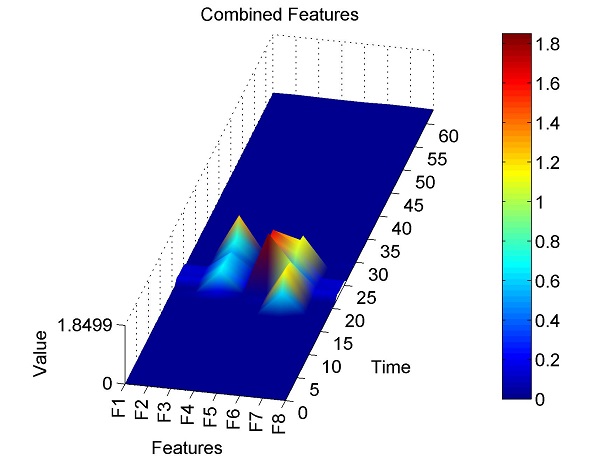} \\
		(a) & (b) & (c)\\
		\includegraphics[height=3cm]{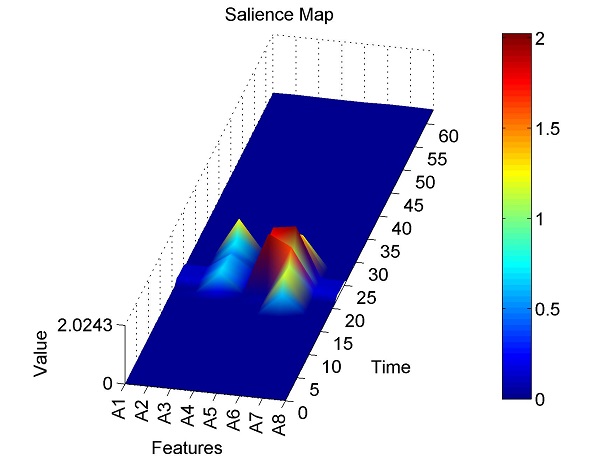}  &
		\includegraphics[height=3cm]{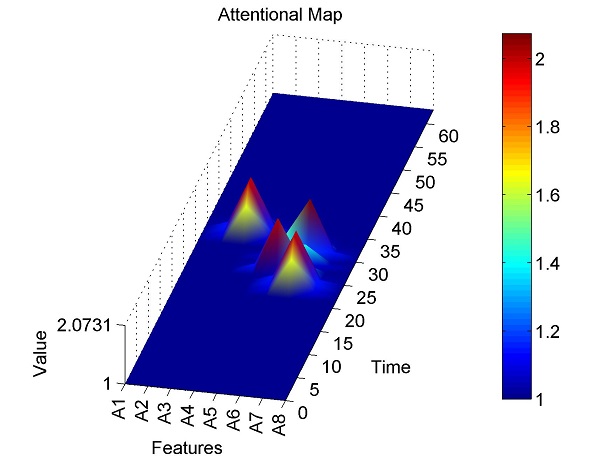} &
		\includegraphics[height=3cm]{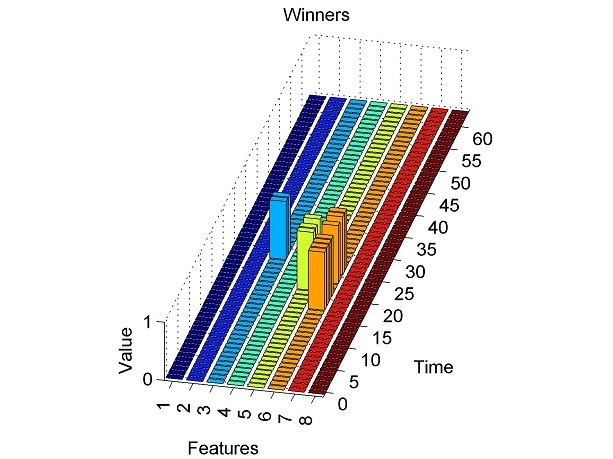}  \\
		(d) & (e) & (f) \\
		\end{tabular}
\end{center}
\caption{Experiment 05: a) $\mathcal{F}_1$ b)  $\mathcal{F}_4$ c) $\mathcal{C}$ d) $\mathcal{L}$ e) $\mathcal{M}$ f) WTA.}
\label{exp05}
   \end{figure}

In this experiment (Figure \ref{exp05}), the top-down dimension used was $\mathcal{F}_4$ with the goal for the attentive robot as: GREATER(0.6), in m/s. The bottom-up feature used was $\mathcal{F}_1$. The results show that the top-down mechanism, that aimed for regions with quite high speed activity, first drove the attention to the stimulus where the speed was inside its goal (tracking the fastest robot), then, in its absence, to those with less movement, but that still generated some activity. However, as all attention was driven by a top-down mechanism, no IOR course took place and the shifting on the attentional focus was promoted only by the change in the speed of the targets and not because a region was suffering from an inhibitory course. It is important to mention that as the bottom-up feature aimed discrepancies in the motion of objects, the moving robots also triggered exogenous processes that helped the course of the endogenous one.  

\section{Conclusions}
\label{sec:conclusion}

This paper investigated the possibility of using sensors with different dimensions to define multiple features for an attentional model. A set of bottom-up and top-down feature mapping functions were designed. The experiments conducted in a high fidelity simulator with mobile robots have demonstrated that the attentional model previously proposed with these feature mapping functions captured some of the key aspects of exogenous and endogenous attentional processes such as: amplification of stimuli, lateral excitation, refraction period and relevant stimulus detection. Moreover, it demonstrated the possibility of composing feature maps through sensors with different characteristics, opening space for combining these traditional sensors with visual ones. Finally, attentional effects such as IOR promotes a natural exploration in the environment. This is particular important if we consider the attentional model as a elicitor of important information to a future decision process. 

\bibliographystyle{splncs}
\bibliography{Thesis}

\end{document}